**Traffic Cameras to detect inland waterway barge traffic: An Application of machine learning.**

**Geoffery Agorku**
Department of Civil Engineering
University of Arkansas, Fayetteville, Arkansas, 72701
gagorku@uark.edu
https://orcid.org/0009-0000-7716-0334

**Sarah Hernandez, PhD**
Department of Civil Engineering
University of Arkansas, Fayetteville, Arkansas, 72701
sarahvh@uark.edu
https://orcid.org/0000-0002-4243-1461

**Maria Falquez**
Department of Civil Engineering
University of Arkansas, Fayetteville, Arkansas, 72701
msfalque@uark.edu

**Subhadipto Poddar, PhD**
Department of Civil Engineering
University of Arkansas, Fayetteville, Arkansas, 72701
sp132@uark.edu
https://orcid.org/ 0000-0003-4211-6688

**Kwadwo Amankwah-Nkyi**
Department of Civil Engineering
University of Arkansas, Fayetteville, Arkansas, 72701
kwadwoa@uark.edu
https://orcid.org/ 0009-0001-1860-2961

Word Count: 7,847 words + 1,500 [6 tables (250 words per table)] = 9,359 words

Submitted 1/5/2024

## ABSTRACT

Inland waterways are critical for freight movement, but limited means exist for monitoring their performance and usage by freight-carrying vessels, e.g., barges. While methods to track vessels, e.g., tug and tow boats, are publicly available through Automatic Identification Systems (AIS), ways to track freight tonnages and commodity flows carried on barges along these critical marine highways are non-existent, especially in real-time settings. This paper develops a method to detect barge traffic on inland waterways using existing traffic cameras with opportune viewing angles. Deep learning models, specifically, You Only Look Once (YOLO), Single Shot MultiBox Detector (SSD), and EfficientDet are employed. The model detects the presence of vessels and/or barges from video and performs a classification (no vessel or barge, vessel without barge, vessel with barge, and barge). A dataset of 331 annotated images was collected from five existing traffic cameras along the Mississippi and Ohio Rivers for model development. YOLOv8 achieves an F1-score of 96%, outperforming YOLOv5, SSD, and EfficientDet models with 86%, 79%, and 77% respectively. Sensitivity analysis was carried out regarding weather conditions (fog and rain) and location (Mississippi and Ohio rivers). A background subtraction technique was used to normalize video images across the various locations for the location sensitivity analysis. This model can be used to detect the presence of barges along river segments, which can be used for anonymous bulk commodity tracking and monitoring. Such data is valuable for long-range transportation planning efforts carried out by public transportation agencies, in addition to operational and maintenance planning conducted by federal agencies such as the US Army Corp of Engineers.

## INTRODUCTION

Barge transportation plays a pivotal role in the logistics industry, especially in countries with extensive river systems like the United States [1]. Barge transportation is considered an environmentally friendly alternative to road and rail transportation, contributing to the reduction of carbon emissions [2]. Barges offer an efficient and cost-effective means for transporting large quantities of goods such as coal, grains, and petroleum products [3]. Compared to other modes of transportation, barge transportation on waterways provides advantages in terms of safety, reliability, and environmental sustainability [4].

Monitoring barge traffic on waterways presents significant challenges as barges, unlike the vessels that tow them, are not typically equipped with tracking devices to monitor their position [5]. In comparison, vessels like tugs and towboats, and other self-propelled vessels operate with Automatic Identification Systems (AIS) tracking equipment. AIS is mandated for safety purposes and assists with navigation [6]. AIS data is shared publicly and can be used to monitor vessel traffic along inland and coastal waterways [7]. Using AIS, vessel traffic (volume), speed, stopping events, and travel times, among other parameters can be measured and monitored to provide performance metrics for our inland waterways. However, barge volumes in terms of number of barges, commodities carried, quantity of commodity, etc. cannot be monitored in real-time (or in near-real time) as the only public source of this information includes the Lock Performance Monitoring System (LPMS) [8], Commodity Flow Survey (CFS), and Waterborne Commerce Statistics Center (WCSC) [9]. Such sources are derived from waybills, surveys, and other passive means. The usage of passive means like waybills and surveys introduce potential inaccuracies. Surveys are inherently limited due to delays in data collection, sampling biases, issues with data quality and response lags. Similarly, waybills are associated with manual data entry errors and delays in reporting. Human errors, intentional or unintentional, and incomplete reporting can compromise the efficacy of an efficient barge monitoring system. All these limitations can lead to potential inaccuracies in the data obtained through these passive means.

Thus, there is value in monitoring barge movements through direct observation and potentially in real-time. Ling et al. [5] developed the TRACC system, a real-time identification and monitoring system for tracking barge-carried hazardous commodities on waterways. This system uses static and trip information databases and tracking devices attached to barges. Key components of the system include an Event Prediction Module, which estimates the position of the barge based on updates from onboard instruments and predicts the arrival of the barge at critical locations. However, this relies on specially installed tracking devices reliant on cell phone connection and solar power. This dependence on solar power and cell signals limits the system's ability for real-time tracking in certain conditions. Also, the prediction model is susceptible to errors in time series forecasting, potential false positives or false negatives in the anomaly detection module, and the reliance on an honor system for barge movement reporting, where operators may miss reporting or omit essential information. Furthermore, the static information database reports on river infrastructure, equipment inventory, and points of interest. The trip information database contains a hierarchy of information related to a barge trip. However, some of these data may change on a longer temporal scale when equipment is bought or sold, or new points of interest are established. This could potentially lead to inconsistencies or inaccuracies in the data.

This study acknowledges these challenges and addresses the need for automated real-time detection of barges and vessels. The novelty of this study lies in being the first to utilize publicly available traffic cameras for detecting barges along inland river systems. Highway traffic cameras (referred to as traffic cameras) are typically installed along interstates or major highways to monitor traffic flows, verify reports of accidents/incidents, and improve response times and operations by connecting to Traffic Management Centers [10]. In locations near inland waterways, bridges, and other critical infrastructure, traffic cameras purposely or opportunistically include a view of the waterway. In these cases, the cameras can effectively monitor marine vessel traffic. For this work, we identified twenty cameras that had views of the Mississippi River, Tennessee River, Ohio River, and Arkansas River systems. Five of these were used to develop the models in this paper to detect barge traffic. Demonstrating the ability to use existing traffic cameras for marine applications may promote the idea of co-usage of camera monitoring and lead to more prolific installation of cameras in marine settings.

The objectives of this study are to:

1. Develop an automated, real-time barge monitoring system that ensures high accuracy in barge detection while being robust to variations in lighting, angles, and perspectives.
2. Evaluate the performance of the developed barge detection system under different conditions, namely fog, rain, and location types, to assess its robustness and reliability in real-world scenarios.

The paper is structured like this: Introduction (topic and motivation), Background (literature review), Methods (methodology and techniques), Results and Discussion (implications and significance), and Conclusion (findings summary and future research suggestions).

## BACKGROUND

This study focuses on barge movements via the US inland waterways, also known as the Marine Highway System. The US inland waterways span approximately 12,000 miles and include commercially active routes for transporting major bulk commodities like grain, coal, and petroleum [16]. These waterways support the movement of about 630 million tons of cargo annually, valued at over $73 billion, which underscores the importance of efficient barge detection systems in maintaining economic vitality [16]. As previously mentioned, there are limited means of capturing real-time movements of barge traffic on inland waterways and as a result there is limited data on which to base performance metrics for planning, operations, and management decisions. Moreover, this becomes increasingly relevant considering the federal government's role in supporting navigation. The U.S. Army Corps of Engineers and the Departments of Transportation, through agencies like the U.S. Coast Guard and the Maritime Administration, play crucial roles in maintaining and operating the navigation system, which includes ensuring the safety and efficiency of inland waterway transport [16]. Automated detection has the potential to produce real-time barge movement data. Because inland waterways serve 38 states and contribute significantly to the nation's economy, with states like Texas and Louisiana shipping over $10 billion worth of cargo annually, the implementation of an advanced barge detection system is not only a technological advancement but a strategic necessity [16].

Recent advancements in computer vision and deep learning techniques transformed the field of automated detection systems across various domains, including transportation [11]–[13]. These techniques empower machines to interpret and comprehend visual information, making them particularly suitable for object detection tasks. While numerous studies explored the application of computer vision and deep learning for object detection in surface transportation [13]–[15] (e.g., roads and rail), a gap exists specifically in the automated detection of barges. Barges are like truck trailers in freight transport applications in that they carry commodities but are non-self-propelled and often not actively tracked. This presents a unique challenge that lends itself to passive detection approaches like vision, Lidar, or radar-based, among other approaches.

With decreasing costs of installation and operation, there is an increasing proliferation of cameras for transportation data collection [18]. Considering freight applications as a key contribution of this study, this section first highlights research on advanced freight detection using passive collection tools, specifically vision-based systems like cameras. The state-of-the-practice and -art approaches for vision detection methods are summarized concurrently with the application areas.

Moghimi et al. [14] proposed a vehicle detection system that uses the Viola-Jones algorithm combined with AdaBoost. The modeled dataset consisted of 576 images of vehicles under different lighting conditions obtained from surveillance videos. The accuracy, completeness, and quality rates of the proposed method were about 94%, 92%, and 87%, respectively in all lighting conditions. Gupta and Gupta [19] designed an automated object detection system for marine environments using cameras. The system employed a support vector machine (SVM) algorithm trained on a dataset of over 2500 images encompassing 16 categories. The accuracy of the proposed method was 67%. The algorithm's performance was compared to the works of Koch et al. [20] and Zhang et al. [21] demonstrating its enhanced results and potential for object detection and classification in coastline surveillance systems. Similarly, Liu et al. [12] developed ship detection methods in river environments using enhanced Convolutional Neural Network (CNN) methods under different weather conditions. A key component of their approach was the implementation of a flexible data augmentation strategy that generated synthetically degraded images to augment the volume and diversity of the original dataset. Their approach significantly outperformed several cutting-edge techniques, including Single Shot MultiBox Detector (SSD), Faster Region Convolutional Neural Network (R-CNN), and You Only Look Once version 3 (YOLOv3), in terms of detection accuracy, robustness, and efficiency. However, like previous studies, it did not focus specifically on barge detection, which is unique given the physical, low-lying structure of the barges themselves. Barges vary significantly in size and configuration. Their appearance in images or video depends on their distance from and angle of the camera [22]. Detecting barges at varying scales and perspectives requires robust object detection algorithms capable of handling these variations [23]. Furthermore, waterways and coastlines often have complex and cluttered backgrounds, with many objects, including boats, buoys, bridges, and buildings [24]. This cluttered environment can hinder the accurate detection of barges, as the system needs to differentiate between the barge of interest and other objects in the scene [25].

Hammedi et al. [26] explored the application of several object detection algorithms, including Faster R-CNN, SSD, and different versions of YOLO, for real-time object detection in river environments. Their dataset consisted of 2,488 images with almost 35,400 annotations. The findings demonstrated that all six algorithms exhibited the ability to detect object classes including riverside (embankment), vessels, persons, infrastructure, and road signals in near real-time. Faster R-CNN achieved the highest accuracy but was slower, while SSD was faster but slightly less accurate, and YOLO was the fastest but had the lowest accuracy. The authors recommended selecting the algorithm based on the specific application requirements, with Faster R-CNN prioritized for accuracy-critical tasks and YOLO for speed-oriented applications. The paper also contributed by providing an annotated dataset for training and adapting deep learning techniques to river environments, showcasing the feasibility of real-time object detection in such environments.

While these studies provided valuable insights into the application of computer vision and deep learning techniques for object detection in transportation, none addressed the challenges of using existing traffic cameras. This is an important challenge as it has implications for the quality of images and viewpoints which affect model performance and transferability. Additionally, much of the existing research targets the detection and classification of self-propelled vessels. Barges are non-self-propelled with physical dimensions (low, flat, grouped, etc.) that make barge detection difficult. These research gaps serve as motivation for the current study, which aims to develop an automated barge detection system that leverages state-of-the-art computer vision and deep learning techniques, using Single Shot Multibox Detector (SSD), EfficientDet, You Only Look Once Version 5 (YOLOv5) and You Only Look Once Version 8 (YOLOv8) models. These models, as indicated in the above review, have shown superior performance in real-time settings, specifically where video feeds are segmented into images, and then fed into feature extraction and classification algorithms.

## METHODS

### Dataset and Data Collection

The dataset for this experiment was collected by monitoring five cameras located at the Cincinnati Covington Bridge (CBB) in Ohio, St. Louis Arch (SLA) in Missouri, Emerson River Bridge (ERB) in Missouri, Mississippi River Bridge (MRB) in Mississippi, and Louisiana River Bridge (LRB) in Mississippi (**Figure 1;** example images in **Figure 3**). These locations were chosen from a cursory review of about twenty cameras located on roadways along inland waterways in the states of Mississippi, Tennessee, Ohio, and Arkansas. The two river locations present diverse river traffic with the Mississippi River locations seeing mostly barges and the Ohio River seeing barges and recreational (riverboat) vessels. The assumed intention of the selected cameras, based on view angles, was for bridge monitoring or tourism/scenic observation. Cameras at ERB and LRB are operated by state transportation agencies and those at CCB, SLA, and MRB are operated by EarthCam [27]. Real-time video feeds were recorded and then used to extract 331 images for annotation (**Figure 2**). For this work, the conversion process was achieved using the OpenCV Python library, a robust computer vision tool.

**Table 1: Image Source and Number of Images in Dataset**

| Image Source / Camera Location | Operating Agency | Number (percentage) of Images in Dataset | Vessel/Barge Traffic (vessels per day[1]) | River Width (Approx) at camera position |
|---|---|---|---|---|
| **ERB**: Emerson River Bridge | Missouri Department of Transportation [28] | 48 (14.4%) | 12 (90% vessels with barges) | 0.4 miles |
| **LRB**: Louisiana River Bridge | | 55 (16.5%) | 13 (79.6% vessels with barges) | 0.6 miles |
| **SLA:** St. Louis Arch | EarthCam (Private) [29] | 60 (18.1%) | 14 (69.3% vessels with barges) | 0.3 miles |
| **MRB:** Mississippi River Bridge | EarthCam (Private) [30] | 142 (43.0%) | 12 (80.8% vessels with barges) | 0.6 miles |
| **CCB**: Cincinnati Covington | EarthCam (Private) [31] | 26 (7.9%) | 5 (45% vessels with barges) | 0.2 miles |
| **Total** | - | 331 | - | - |

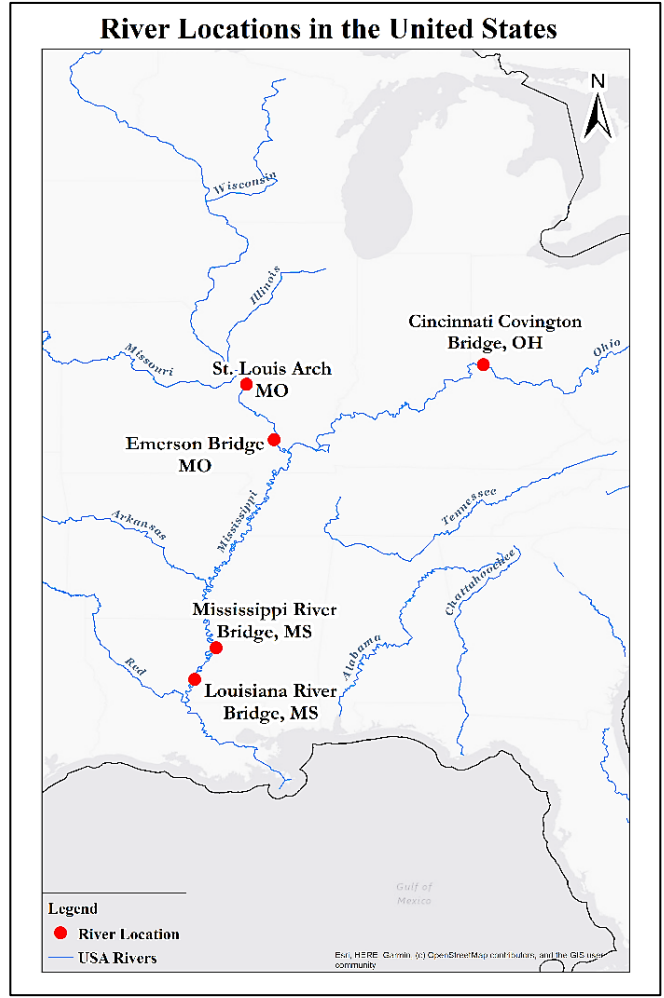

**Figure 1: Locations of traffic cameras used for data collection for model development and testing.**

Notes: 1. 'Day' refers to daytime hours only.

To ensure generalization of the model, images were collected during different times of the day (day and night) and different weather conditions (clear, rainy, and foggy weather) (**Figures 2 & 3**). Data was collected between 9th October 2022 and 3rd May 2023. Background images are images with no objects added to a dataset to reduce False Positives (FP). Twelve percent of the images in the dataset were background images [32].

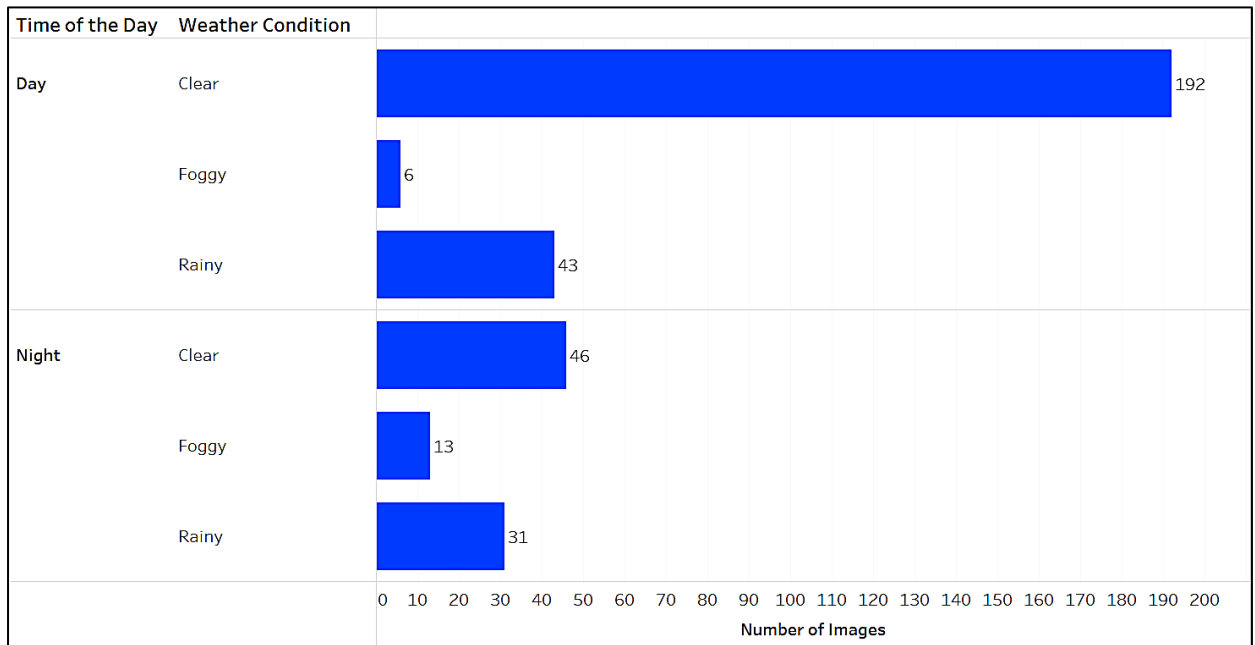

**Figure 2: Distribution of Samples in the Dataset by Time of Day and Weather Conditions**

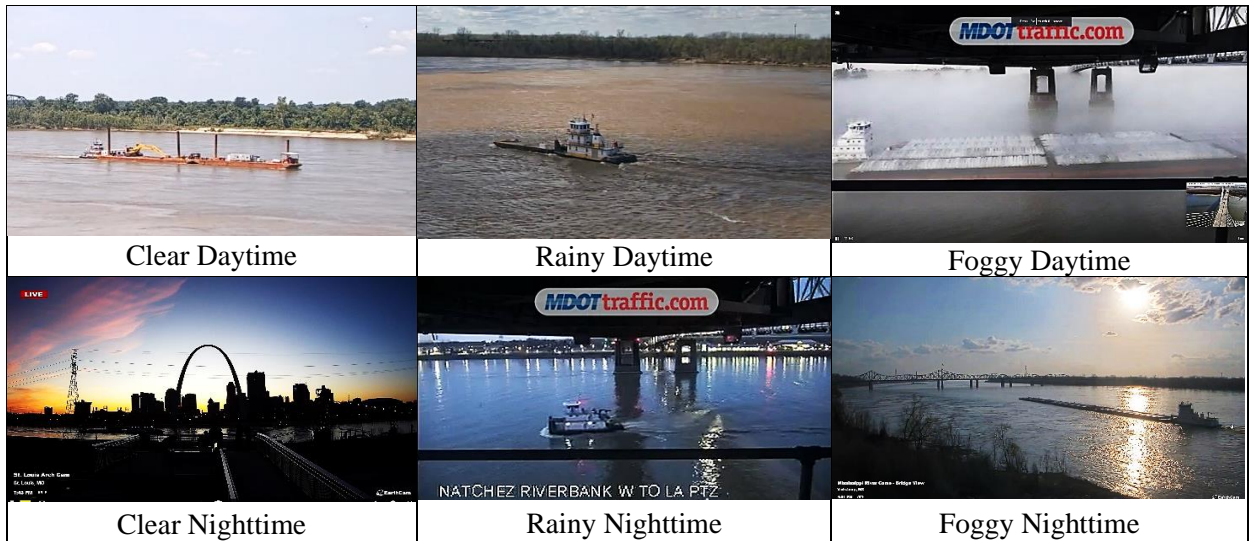

**Figure 3: Dataset instances showing diverse environmental and weather conditions.**

**Image Annotation and Data Augmentation**

There are no publicly available annotated datasets containing vessel and barge images. Thus, the annotated images were generated specifically for this study from the five camera locations. Annotation involved drawing rectangular bounding boxes around vessels and barges (**Figure 4**). A consensus-based approach was employed by a team of annotators to resolve discrepancies and ensure accurate annotation. The manual annotation process was facilitated using the Computer Vision Annotation Tool (CVAT) [33] to draw precise bounding boxes around barges and vessels, minimizing annotation errors and enhancing annotation quality. CVAT was used because it provided a user-friendly web-based interface, which simplified the annotation process. Researchers can utilize this dataset as a benchmark for future studies. The annotated dataset consisted of a total of 331 images.

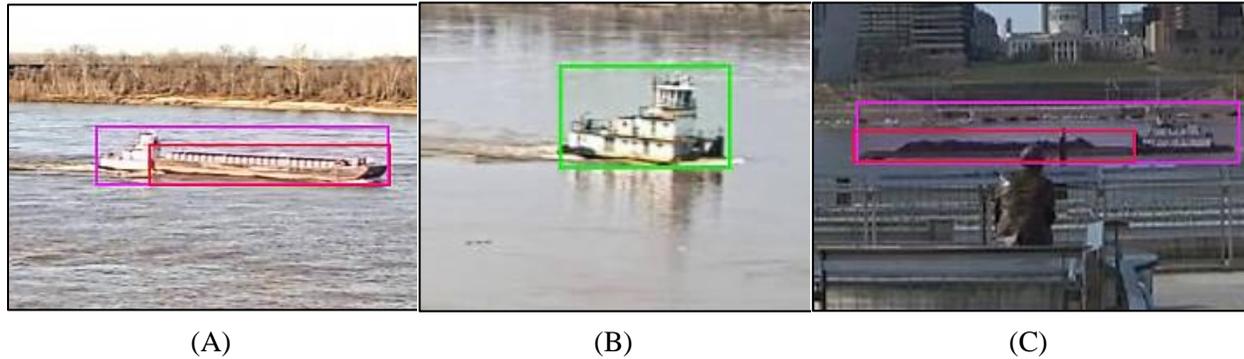

(A)                                (B)                                (C)

**Figure 4: Sample annotations of Barges and Vessels on waterways. (A) and (C) denotes vessels with barges, (B) denotes a vessel without a barge.**

To address the challenges posed by adverse weather conditions and low illumination in real-world imaging scenarios, data augmentation techniques were employed [12] to expand the data set with 441 additional images generated using label-preserving transformations with Roboflow [34]. The augmentation techniques included random cropping [35], Gaussian blur, horizontal flipping [36], scaling, rotation, shear, saturation adjustment, brightness adjustment, exposure adjustment, cutout, and the addition of noise to the images (**Figure 5**). Random cropping generated multiple sub-images from each original image, allowing the model to learn from diverse perspectives and improve its object detection capabilities [37]. Gaussian blur simulated adverse weather conditions, enabling the model to better handle degraded image quality. Horizontal flipping introduced additional variety and helped the model generalize better to different barge orientations [38]. Simulating variations in lighting conditions, angles, and perspectives through scaling, rotation, shear, and other transformations enhanced the model's barge detection accuracy under diverse scenarios. However, it is essential to acknowledge that data augmentation techniques have limitations. Excessive application of certain augmentation methods may introduce artifacts or distortions that could impact the model's performance. Nevertheless, the augmented dataset, comprising both original images and their augmented versions, offered a more diverse and representative training set.

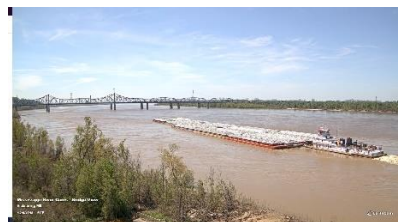

(A) Original Image

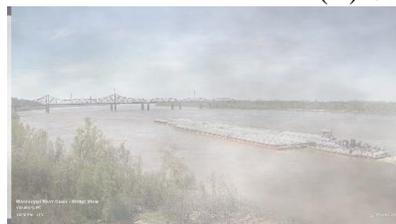           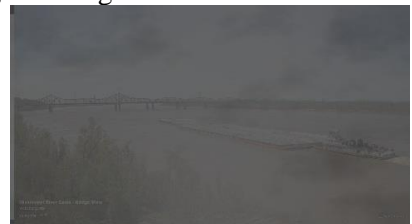

(B) Foggy daytime augmentation  (C) Foggy nighttime augmentation

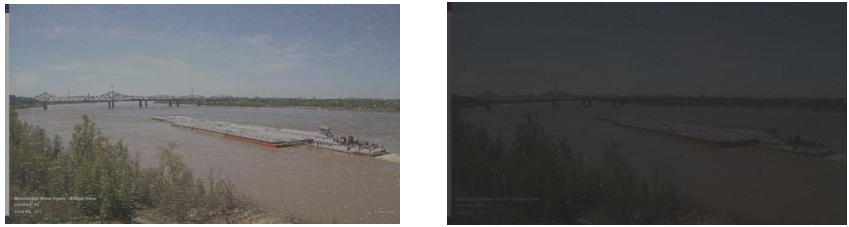

(D) Rainy daytime augmentation (E) Rainy nighttime augmentation

**Figure 5: Sample images before and after applying augmentations.**

## Model Selection and Transfer Learning

The study employed state-of-the-art Convolutional Neural Networks (CNNs) for barge detection. CNNs are very effective for image classification and recognition tasks [39]. A CNN consists of several building blocks: convolutional layers, pooling layers, and fully connected layers [39].

1. Convolutional Layers: The convolutional layers form the foundation of a CNN. They scan the input image and build feature maps using a set of learnable filters (also known as kernels or weights) [40]. Each filter performs a mathematical operation (convolution) on a small portion of the input image and outputs a single value [40]. All filter output values are arranged in a 2D feature map. Convolutional layers help in the detection of low-level characteristics like edges and corners [40].

2. Pooling Layers: The pooling layers reduce the feature maps produced by the convolutional layers [41]. The feature maps are down-sampled by obtaining the maximum or average value of a small portion of the feature map [41]. This reduces the network's computational complexity and makes it more robust to translation and rotation of the input picture [41].

3. Fully Connected Layers: They classify the input image into various classes [42]. They take the preceding layer's flattened output (the feature vector) and apply a set of learnable weights to generate a probability distribution across the classes [42]. To generate the final output probabilities, the output of the fully connected layers is fed into a SoftMax activation function [43].

A CNN's architecture generally comprises multiple convolutional and pooling layers followed by one or more fully connected layers (**Figure 6**) [43]**.** Hyperparameters such as filter size, stride, and padding are tweaked for optimal performance [43]. During training, the network learns the optimal values of the filters and weights by minimizing a loss function (such as cross-entropy loss) using backpropagation [44]. The backpropagation algorithm computes the gradients of the loss with respect to the parameters of the network and updates them using an optimization algorithm (such as stochastic gradient descent) to minimize the loss [44].

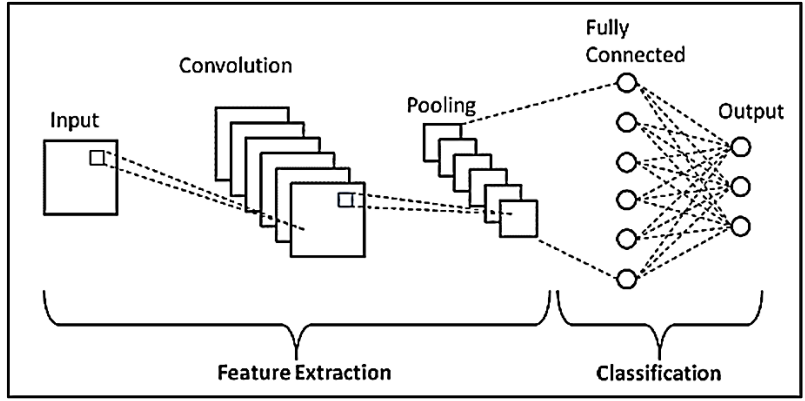

**Figure 6: Schematic of the CNN Architecture [45]**

For this work, SSD [46], EfficientDet [47], YOLOv5 [48], and YOLOv8 [49], four widely recognized pre-trained CNN models, were chosen based on their architectural strengths in object detection and their reported successes in related research domains [12], [26], [50]. SSD was chosen for its speed, efficiency, accuracy, and robustness [51]. EfficientDet is recognized for effectively balancing the tradeoff between speed and accuracy within a cost-effective parameter space [52]. YOLOv5, the predecessor of YOLOv8, was selected for its advanced user-friendliness, making it an excellent choice for projects where simplified setup and ease of use are essential. YOLOv5's architecture and design focus on delivering an accessible tool for a wide range of applications. YOLOv8 is preferred when speed and accuracy are paramount. It excels in real-time object detection scenarios, such as this barge detection application, where rapid decision-making based on high-accuracy object detection is vital. YOLOv8 builds upon YOLOv5 and achieves this by implementing optimizations in its architecture that prioritize speed and maintain high levels of accuracy, making it ideal for real-time use cases.

Transfer learning was used to adapt the pre-trained models to the unique objective of barge detection [53]. This resulted in less training time and higher detection accuracy [53]. This involved fine-tuning specific layers of the models on our annotated dataset. By focusing the learning process on the relevant features for barge detection, transfer learning enabled us to leverage the models' pre-existing knowledge while tailoring them to our specific task of real-time barge detection.

**Hyperparameter Tuning**

The training process for the YOLO models requires tuning of approximately 30 hyperparameters [54] (**Table 2**) including network architecture, optimization algorithms, and regularization techniques. The tuning process was conducted through a systematic, iterative approach. Specific ranges and step sizes were defined for each hyperparameter, and various combinations were evaluated to gauge their impact on the model's performance. Performance evaluation was carried out using the F1-score (**Eq. 1-3**) [55]. Overall, hyperparameter tuning increased the model's performance from an F1 score of 84% to 96%. Future work will explore the use of genetic algorithms and Markov processes or Bayesian optimization algorithms to determine optimal hyperparameters.

$$\text{F1 score} = \frac{2 \times \text{Precision} \times \text{Recall}}{\text{Precision} + \text{Recall}} \text{ [55]} \qquad \textbf{Equation 1}$$

Where,

$$\text{Precision} = \frac{\text{True Positives}}{\text{True Positives} + \text{False Positives}} \text{ [56]} \qquad \textbf{Equation 2}$$

$$\text{Recall} = \frac{\text{True Positives}}{\text{True Positives} + \text{False Negatives}} \ [56] \qquad \textbf{Equation 3}$$

**Table 2: Hyperparameter value ranges for training YOLOV5 and YOLOV8 models**

| Hyperpara meter | Description | Tuning Range | Selected Values | | | |
|---|---|---|---|---|---|---|
| | | | YOLO v5 | YOLO v8 | SSD | EfficientDet |
| Batch size | Number of images in each training batch | 4-32 | 8 | 4 | 4 | 4 |
| Epochs | Number of times the entire dataset is passed through the model | 200-1500 | 600 | 600 | 1000 | 500 |
| Image Size | Size of input images | 320, 512, 640, 896, 1024, 1216 | 896 | 1216 | 320 | 512 |
| Optimizer | Algorithm used to update model weights | Adam, SGD, AdamW, RMSProp, Momentum | Adam | SGD | Adam | Momentum |
| Momentum | Amount of contribution from previous weight updates | 0.9-0.99 | 0.937 | 0.937 | 0.9 | 0.9 |
| Weight Decay | Regularization term to prevent overfitting | $0.00001 - 0.001$ | 0.0005 | 0.0005 | 0.00004 | 0.000004 |
| IOU | Ratio of overlap to total area | $0.2 - 0.7$ | 0.2 | 0.7 | 0.4 | 0.5 |
| Lr0 | Initial learning rate | 0.01-0.00126 | 0.01 | 0.01 | 0.10 | 0.08 |

**Model Training and Validation**

A stratified random sampling approach was used to divide the data into training, validation, and testing sets. The training set optimized model parameters, while the validation set fine-tuned hyperparameters and monitored performance. The testing dataset remained separate for final model evaluation. To prevent overfitting during the training process, early stopping was implemented using the patience parameter [57]. The patience value used was 100. This technique allowed for the termination of training when the model's performance did not improve beyond the specified number of epochs. The best-performing model was selected for further assessment on the testing set. In total, 771 images were divided into training, validation, and testing sets with a ratio of 70:15:15 (**Figure 7**). Only original images (not augmented) were used in the testing dataset. Later sensitivity training divided the dataset by location and image condition (rain, fog, etc.).

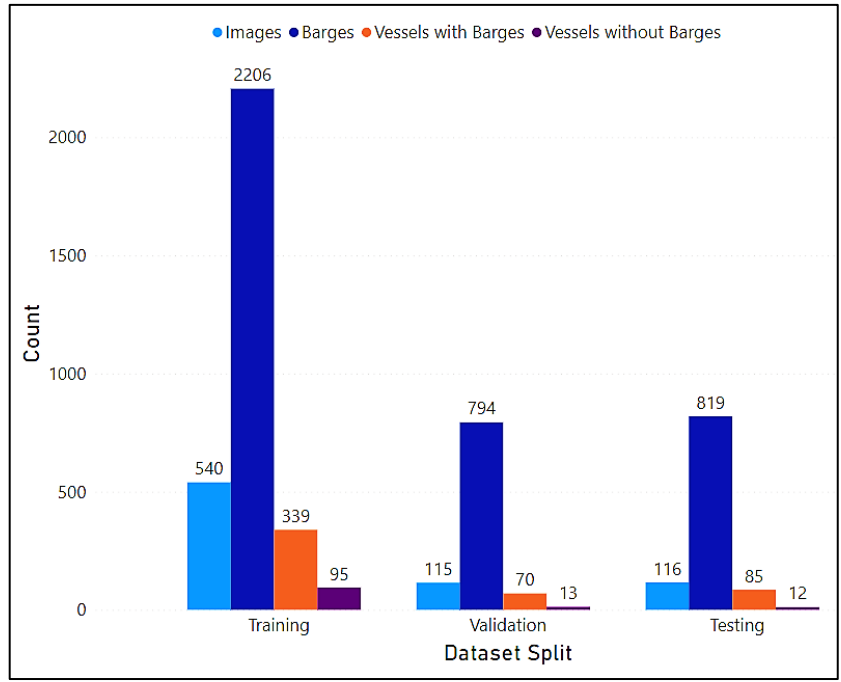

**Figure 7: Summary of Dataset Split and Vessel Characteristics**

**Spatial Transferability Analysis Framework**

Spatial transferability is the generalizability of the model to different environments [58, 45] and enables identification of potential challenges and development of strategies to enhance transferability [60]. In the context of this paper, that refers to different inland waterways, view angles, lighting conditions, and vessel/barge traffic mixes **(Table 3).** The spatial transferability was performed using cross-validation. A model was trained on four of the five locations and tested on the fifth location.

**Table 3: Environmental Characteristics of Data Collection Locations**

| Location | Lighting | Water Coloration | Occlusion levels | Percent of river area in image |
|---|---|---|---|---|
| **ERB**: Emerson River Bridge | Clear and foggy conditions present | Dark brown, muddy, murky | Heavy occlusion | River comprised approximately 80% of the image's viewable area |
| **LRB**: Louisiana River Bridge | Varies between clear and foggy weather | Dark brown | Partial occlusion | 80% |
| **SLA:** St. Louis Arch | Mostly clear, good lightening | Dark brown | Partial Occlusion | 20% |
| **MRB:** Mississippi River Bridge | Variation between clear and foggy weather | Dark brown; muddy and murky during rainy conditions | No occlusion | 90% |
| **CCB**: Cincinnati Covington | Generally clearer than Mississippi | Greenish blue | Partial occlusion | 60% |

## RESULTS AND DISCUSSION

The classification was based on a five-class scheme **(Figure 8)**:

A. **No Detection (no vessel, no barge)**: No vessel or barge(s) detected in the image. **(Figure 8A)**
B. **Vessel Detected without Barge, No Barge Detected**: The model detects a vessel, but there is no barge present **(Figure 8B)**
C. **Vessel Detected without Barge, Barge Detected:** The model detects a vessel without barge(s), but there is also a barge in the image **(Figure 8C).**
D. **Vessel Detected with Barge, Barge Detected**: The model detects both the vessel towing barge(s) and the barge(s) present in the image **(Figure 8D).**
E. **Barge Detected**: The model detects a barge, but there is no vessel present **(Figure 8E).**

The model occasionally introduces a sixth class denoted as **"(F) Vessel Detected with Barge, No Barge Detected**," alongside the five target classes. This additional class emerges when the model mistakenly identifies or predicts a vessel with a barge even when no barge is actually present. This misclassification is likely attributed to a consistent association where the presence of a barge is frequently linked to its being pushed by a tug or tow boat. This scenario is reflected in the confusion matrix presented in Table 5.

There are also instances in the image dataset where barges are present, but no vessel is present due to the position of the barge/vessel pair within the view of the image **(Figure 8E)**. Images may contain more than one vessel and/or more than one set of barges, although this is rare (approximately 2% of the annotated data) **(Figure 8C).**

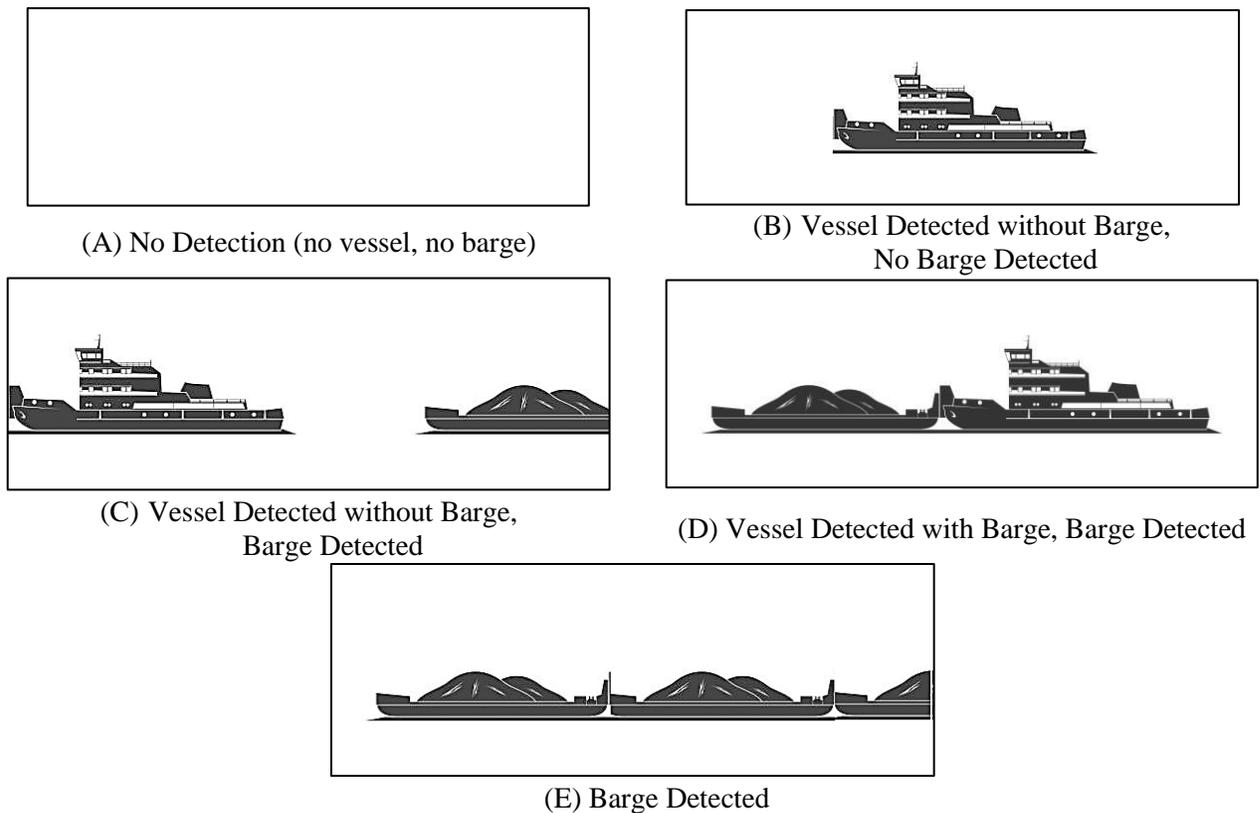

(A) No Detection (no vessel, no barge)

(B) Vessel Detected without Barge, No Barge Detected

(C) Vessel Detected without Barge, Barge Detected

(D) Vessel Detected with Barge, Barge Detected

(E) Barge Detected

**Figure 8: Schematic representation of the five-class scheme**

On the testing dataset which contains 15% (116 images) of the sample images across all five locations, the YOLOv8 achieved an F1 score of 96% at a speed of 34 frames per second (fps), YOLOv5 achieved 86% at a speed of 32 fps, SSD achieved an F1 score of 79% at a speed of 42 frames per second and EfficientDet achieved an F1 score of 77% at 33 frames per second (**Table 4**). The YOLOv8 model demonstrated class-specific F1 scores of 100% for (A) No Detection (no vessel, no barge) Class, 100% for (B) Vessel Detected without Barge, No Barge Detected Class, 100% for (C) Vessel Detected without Barge, Barge Detected Class, 93% for (D) Vessel Detected with Barge, Barge Detected Class and 85% for (E) Barge Detected Class (**Table 4**). Comparing the class F1 scores and speed between models, the YOLOv8 model consistently outperformed all the other three models. The models were carried out on Google Colab, leveraging the computational capabilities of the NVIDIA Tesla V100 GPU equipped with 8 GB of RAM.

**Table 4: Performance Comparison of Object Detection Models YOLOv5 and YOLOv8**

| Model | Class F1-scores (%) | | | | | | Speed (fps) | F1 Score (%) |
|---|---|---|---|---|---|---|---|---|
| | A | B | C | D | E | F[1] | | |
| EfficientDet | 73.3 | 57.1 | 100 | 77.6 | 79.3 | - | 33.1 | 77.4 |
| SSD | 78.6 | 60 | 100 | 77.6 | 79.3 | - | 42.3 | 79.1 |
| YOLOv5 | 96.3 | 87 | 100 | 89.1 | 55.2 | - | 31.9 | 85.5 |
| YOLOv8 | 100 | 100 | 100 | 92.6 | 85 | - | 34.0 | 95.5 |
| No. of Samples per class | 13 | 12 | 1 | 70 | 20 | 0 | - | 116 |

[1.] Classes not present in the original dataset but falsely predicted by the YOLO models.

**Table 5: Cross Classification Matrix for the YOLOv8 Model**

| Count | | Predicted | | | | | | Total (Obs.) | Accuracy |
|---|---|---|---|---|---|---|---|---|---|
| | | A | B | C | D | E | F[1] | | |
| Observed | A | 13 | 0 | 0 | 0 | 0 | 0 | 13 | 100% |
| | B | 0 | 12 | 0 | 0 | 0 | 0 | 12 | 100% |
| | C | 0 | 0 | 1 | 0 | 0 | 0 | 1 | 100% |
| | D | 0 | 0 | 0 | 63 | 3 | 4 | 70 | 90% |
| | E | 0 | 0 | 0 | 3 | 17 | 0 | 20 | 85% |
| | F[1] | 0 | 0 | 0 | 0 | 0 | 0 | 0 | - |
| Total (Predicted) | | 13 | 12 | 1 | 66 | 20 | 4 | 116 | 96% |

[1] Classes not present in the original dataset but falsely predicted by the YOLO models.

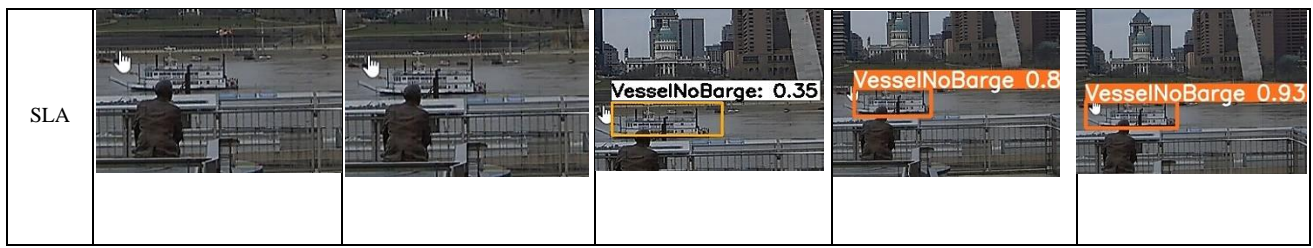

| Location | (A) Original image | (E) EfficientDet | (D) SSD | (B) YOLOv5 | (C) YOLOv8 |
|---|---|---|---|---|---|
| MRB | | VesselWithBarge: 0.56 | 0.51 | VesselBarge 0.87 0.74 | VesselWithBarge 0.77 |
| LRB | | VesselNoBarge: 0.14 | VesselNoBarge: 0.23 | VesselNoBarge 0.76 | VesselNoBarge |
| ERB | | | | Barge 0.85 | |

**Figure 9: Sample detections of models at different locations**

## Spatial Transferability

A spatial transferability analysis evaluated the model's generalization capabilities on unseen locations. The model was trained on four locations along the Mississippi River (ERB, SLA, MRB, LRB) and tested on an unseen location along the Ohio River (CCB). The selection of the CCB location as the holdout (test) dataset demonstrates the following model capabilities. First, the CCB location offers geographical diversity compared to the training locations along the Mississippi River. The Ohio River exhibits unique characteristics, including different water flow patterns, depths, and widths. Second, the CCB location presents operational challenges that differ from the locations along the Mississippi River. Factors such as environmental conditions, lighting, camera angles, water color, and the physical characteristics of the CCB bridge itself differ from those encountered in the Mississippi River settings (**Figure 9**). There were 747 images in the training dataset which includes augmented images and 26 images in the test dataset which does not contain augmented images. The same hyperparameters determined using the original training/validation data set (e.g., all five sites) are applied.

An F1-score of 97.2% resulted for the CCB location as the holdout test set with class-specific accuracies of 100% for (A) No Detection (no vessel, no barge) class, 100% for the (B) Vessel Detected without Barge, No Barge Detected class and 91.7% for the (D) Vessel Detected with Barge, Barge Detected class. It is important to note that the testing dataset did not include class C and E images because these categories are not present in the dataset for the CCB location.

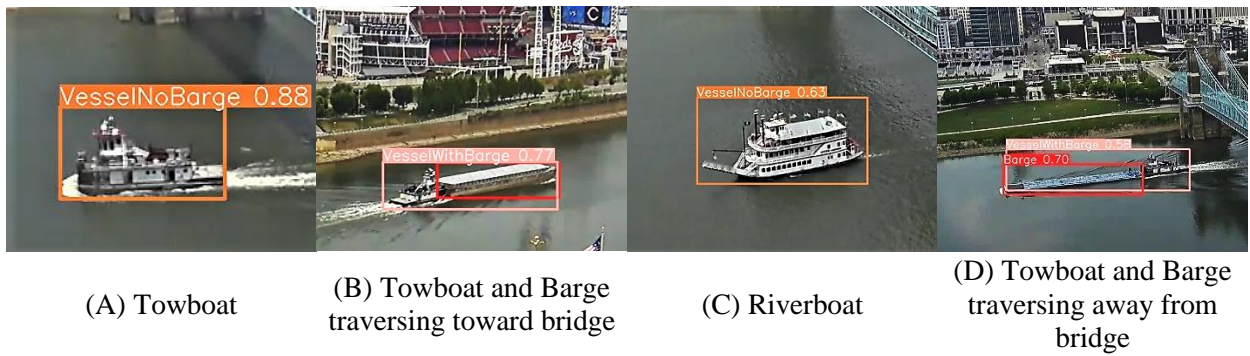

| (A) Towboat | (B) Towboat and Barge traversing toward bridge | (C) Riverboat | (D) Towboat and Barge traversing away from bridge |
|---|---|---|---|

**Figure 9: Sample detections from spatial transferability analysis on CCB**

**Sensitivity to Environmental Characteristics**

Sensitivity analysis was performed to evaluate the robustness of the model under different weather and environmental conditions: rain (**Figure 10A-C**) and fog (**Figure 10D-F**).

**Rain Sensitivity Analysis**

For the rain sensitivity analysis, the model was trained without rain images or rain augmentation and then tested on rainy images. Hyperparameters learned on the full model (all five locations for testing and training) were applied to the models in this section. The model trained without rain train images or rain augmentation achieved an F1 score of 82.3% during training and a score of 90.8% on the test dataset. For the test dataset with the presence of rain conditions (74 samples), class-specific accuracies were as follows.

- (A) No Detection (no vessel, no barge): 88.9%
- (B) Vessel Detected without Barge, No Barge Detected class: 100%
- (D) Vessel Detected with Barge, Barge Detected class: 94.3%
- (E) Barge Detected class: 80%

It is important to note that the testing dataset did not include images from class C as these were not observed under rain conditions at any of the locations.

**Fog Sensitivity Analysis**

For fog conditions, the model was trained without foggy images or fog augmentation and then tested on foggy images. The model trained without foggy images or fog augmentation achieved an F1 score of 80.8% during training and a score of 81.9% on the test dataset. For the test dataset with the presence of fog conditions (19 samples), class-specific accuracies were as follows.

- (A) No Detection (no vessel, no barge): 100%
- (B) Vessel Detected without Barge, No Barge Detected class: 100%
- (D) Vessel Detected with Barge, Barge Detected class: 77.8%
- (E) Barge Detected class: 50%

It is important to note that the testing dataset did not include images from class C as these were not observed under fog conditions at any of the locations.

**Table 6: Model performance under rain and fog sensitivity conditions**

| Condition | Class F1-scores (%) | | | | | | No. test samples | F1 Score (%) |
|---|---|---|---|---|---|---|---|---|
| | A | B | C | D | E | F[1] | | |
| Fog | 88.9 | 100 | - | 94.3 | 80 | - | 74 | 90.8 |
| Rain | 100 | 100 | - | 77.8 | 50 | - | 19 | 81.9 |

[1.] Classes not present in the original dataset but falsely predicted by the YOLO models.

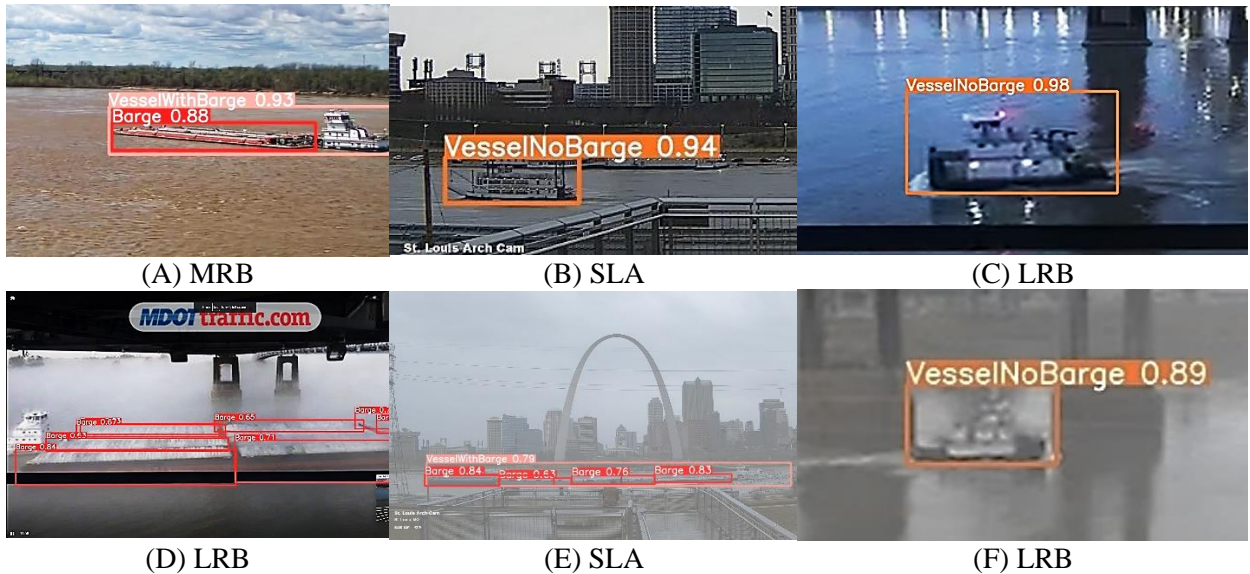

(A) MRB             (B) SLA             (C) LRB

(D) LRB             (E) SLA             (F) LRB

**Figure 10: Sample detections for sensitivity analysis depicting rainy conditions (A)-(C) and foggy conditions (D)-(F)**

## CONCLUSION

This paper presents an object detection model for vessels and barges on inland waterways. The novelty of the approach is twofold. First, the study uses existing highway traffic cameras rather than marine-specific cameras. Using opportune view angles, images of river traffic can be captured. This presents an opportunity to leverage existing investments in highway detection infrastructure for multimodal traffic detection, e.g., roads and waterways. Second, while datasets such as the AIS are available for real-time marine vessel tracking, this data tracks self-propelled vessels like tugs and tow boats and does not track the commodity-carrying entity, i.e., the barge. Barge movements indicate where commodity tonnages are traveling and transferring. Thus, by detecting barges in traffic camera images, data on commodity movements may be more readily available.

The study evaluates four CNN models: SSD, EfficientDet, YOLOv5, and YOLOv8, using a dataset of 771 annotated images collected from five, real-time traffic cameras positioned along the Mississippi and Ohio Rivers. Images are extracted from video feeds. The YOLOv8 model achieves an overall F1-score of 96% at a speed of 34 fps with class-specific accuracies above 85% for all five classes. The YOLOv8 model outperforms YOLOv5 which achieved an F1-score of 86% at a speed of 32 fps, the SSD model which achieved an F1 score of 79% at 42 frames per second, and the EfficientDet model which achieved an F1 score of 77% at 33 frames per second. The overall best model was the YOLOv8 model whose worst performing class ((E) Barge Detected Class), had an F1-score was the with a score of 85% with all wrong instances misclassified as (D) Vessel Detected with Barge, Barge Detected. This may be attributed to the model learning to associate barges with vessels. As a result, when faced with isolated barge images without vessels, the model might mistakenly classify them as if they were towed by vessels due to the shared visual features between the two scenarios. The processing speed was 34 frames per second (fps), resulting in a total runtime of 3.41 seconds for the analysis of 116 test images. This observation signifies the model's efficiency within the specified experimental setup. However, the ultimate viability of the model for industrial implementation may rest upon further optimization and targeted assessments of specific requirements.

Spatial transferability and robustness to environmental conditions including rain and fog was assessed through cross-validation. For spatial transferability, a model was trained on images from cameras along the Mississippi River and tested on images from the Ohio River. Results show an F1-score of 97% for the Ohio River location as the holdout test set with class-specific accuracies above 92% for all classes. Two models were trained without fog or rain images and tested on fog and rain images respectively. The results show the model trained without rain images or rain augmentation achieved an F1 score of 82% and a score of 91% on the test dataset. For fog conditions, the model was trained without foggy images or foggy augmentation and then tested on foggy images. The model trained without foggy images or foggy augmentation achieved an F1 score of 81% and a score of 82% on the test dataset.

Although data for model development and testing was drawn from five traffic cameras that covered two different river systems and multiple view angles, etc., the data can be expanded in future studies to further improve the generalizability of the models. Future research should focus on expanding the dataset by including data from additional cameras, diverse waterway contexts, and more view angles and evaluating the models' performance in diverse waterway contexts to ensure their applicability in real-world maritime settings. Furthermore, future work could also concentrate on improving model performance, particularly in scenarios where the camera's viewing angle does not align with the path of the barge. This could involve refining the model's ability to detect isolated barges accurately.

The approach outlined in this paper makes strategic use of traffic cameras for barge detection. This is a novel use of an existing resource though use of traffic cameras comes with some constraints. Publicly available cameras do not cover all inland waterways. However, a cursory search was conducted and twenty cameras with views of the Mississippi River, Tennessee River, Ohio River, and Arkansas River systems were identified. Another constraint is the dependency on the camera aligning with the path of the barge. In scenarios where the camera's viewing angle does not align with the path of the barge or location of interest, the effectiveness of monitoring may be reduced. Lastly, some many consider the repurposed use of traffic cameras for waterways monitoring to have privacy and ethical implications. In the context of our paper, we argue that traffic cameras routinely collect vehicle-related data without explicit permission from drivers, and this practice has become a standard for traffic monitoring. As such, we don't perceive inherent ethical concerns in the collection of marine traffic data using a similar approach. However, it's essential to recognize that ethical concerns may indeed arise when data collected for one specific purpose, such as traffic monitoring, is repurposed for another, like marine monitoring, without obtaining appropriate permissions or considering the privacy implications. Future work can delve deeper into these concerns and explore potential solutions or guidelines to address privacy and ethical implications effectively. Despite this constraint, our findings offer a starting point for the optimization and placement of cameras to maximize their effectiveness as a future advancement of this work. This practical challenge emphasizes the need for strategic camera positioning to ensure comprehensive monitoring of waterways.

Another practical implication of the study lies in the potential integration of vessel tracking data, such as AIS, with the automated barge detection system. Future advancements in this integration have the potential to offer a comprehensive approach to barge detection. By combining object detection models with AIS data, the system can improve real-time commodity tracking, making it feasible to track the movement of specific commodities. This is invaluable for supply chain management, trade, and logistics, as it allows for real-time visibility into the transportation of goods via waterways. Furthermore, this integration can enhance situational awareness and decision-making, leading to improved safety. By providing real-time tracking and monitoring of barge traffic, transportation agencies and federal agencies such as the US Army Corp of Engineers can make informed decisions about long-range transportation planning, operational and maintenance planning, and freight movement data. This could lead to cost savings, improved resource allocation, and increased productivity in the transportation sector. Future work will examine case studies that specifically delve into and quantify the potential cost savings and economic benefits of implementing our model on a broader scale. While this study represents an intermediate approach toward ultimately using AIS data for barge traffic monitoring, it emphasizes the need for comprehensive monitoring and data management.

## ACKNOWLEDGMENTS

The authors acknowledge the support and sponsorship provided by the US Army Corps of Engineers in conducting this research.

## AUTHOR CONTRIBUTIONS

The authors confirm their contributions to the paper as follows: study conception and design: Geoffery Agorku, Sarah Hernandez; data collection: Geoffery Agorku, Maria Falquez; analysis and interpretation of results: Geoffery Agorku, Sarah Hernandez, Subhadipto Poddar; draft manuscript preparation: Geoffery Agorku, Subhadipto Poddar; draft manuscript preparation: Sarah Hernandez, Subhadipto Poddar, Kwadwo Amankwah-Nkyi. All authors reviewed the results and approved the final version of the manuscript.